\documentclass[11pt]{article}

\usepackage[]{acl}

\usepackage{times}
\usepackage{latexsym}

\usepackage[T1]{fontenc}

\usepackage[utf8]{inputenc}

\usepackage{microtype}

\usepackage{microtype}
\usepackage{graphicx}
\usepackage{subfigure}
\usepackage{booktabs} 
\usepackage{url}
\usepackage{amsmath}
\usepackage{array}
\usepackage{adjustbox}
\usepackage{multirow}
\usepackage{longtable}
\usepackage{enumitem}

\title{STRUM-LLM: Attributed and Structured Contrastive Summarization}

\author{
  Beliz Gunel, James B. Wendt, Jing Xie, Yichao Zhou, Nguyen Vo, Zachary Fisher, Sandeep Tata \\
\texttt{\{bgunel,jwendt,lucyxie,yichaojoey,nguyenvo,zachfisher,tata\}@google.com}\\
  Google Research, Mountain View, CA, USA\\
}

\begin{document}
\maketitle
\begin{abstract}
Users often struggle with decision-making between two options (A vs B), as it usually requires time-consuming research across multiple web pages. We propose STRUM-LLM that addresses this challenge by generating attributed, structured, and helpful contrastive summaries that highlight key differences between the two options. STRUM-LLM identifies \textit{helpful contrast} –- the specific attributes along which the two options differ significantly and which are most likely to influence the user's decision. Our technique is domain-agnostic, and does not require any human-labeled data or fixed attribute list as supervision. STRUM-LLM attributes all extractions back to the input sources along with textual evidence, and it does not have a limit on the length of input sources that it can process. STRUM-LLM Distilled has 100x more throughput than the models with comparable performance while being 10x smaller. In this paper, we provide extensive evaluations for our method and lay out future directions for our currently deployed system.

\end{abstract}

\section{Introduction}

Today, decision-making is often hindered by information overload, and we need tools that can efficiently and effectively distill key differences between options. We introduce STRUM-LLM, a large language model (LLM) based system that aids users with their \textit{A vs B} decisions such as \textit{iPad vs Microsoft Surface}, \textit{Mammogram vs Ultrasound}, or \textit{Lowe's vs Home Depot} by providing them with structured contrastive summaries that highlight the helpful contrast between the two options. Our system is currently deployed in an LLM-enhanced web-search product serving millions of users. Our approach is domain-agnostic, does not require any human labeled data or fixed attribute list as supervision, and can process arbitrarily long input sources in an attributable way.

\begin{figure}[h!]
    \centering
    \includegraphics[width=\linewidth]{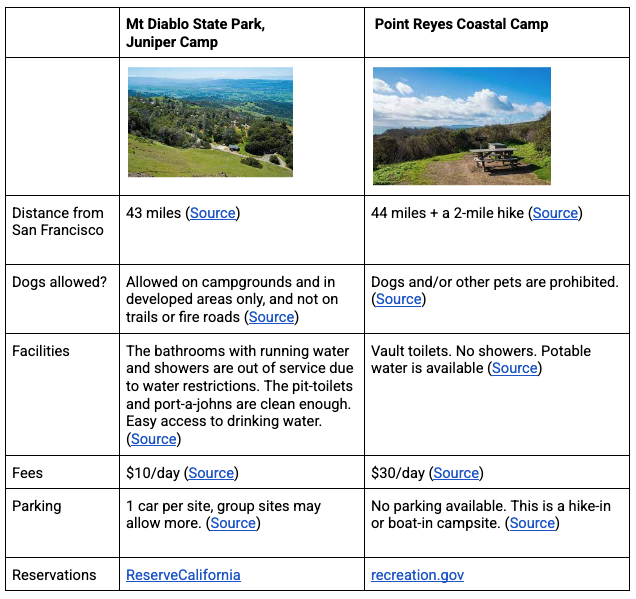}
    \caption{STRUM-LLM aims to produce an \textbf{attributed} (grounded in the input sources), \textbf{faceted} (a row per attribute), and \textbf{helpful} (relevant and contrastive attributes) summary for an A vs B comparison.}
    \label{fig:campsite_mock}
\end{figure}

At the core of STRUM-LLM is the concept of a \textit{helpful comparison}, which is guided by the following desiderata. First, it prioritizes clear attribution to sources, ensuring all information is traceable. High-contrast and important attributes between two options are identified to aid discernment. Second, shown values are consistent, non-redundant, and reflect the majority opinion. Finally, attributes are ranked to highlight the most relevant and contrasting aspects. We present the final summary in a \textit{faceted} way, showing a row per attribute as in Figure~\ref{fig:campsite_mock}. Overall, our contributions that led to a real-world deployed system are in the following:
\begin{itemize}
    \item \textbf{Desiderata and Novel Evaluation Metrics for a \textit{Helpful Comparison}}: We define desiderata and a set of evaluation metrics for identifying a \textit{helpful comparison} and measuring the performance of models that are designed for that task. We demonstrate our automated evaluation metrics correlate well to human judgement in Table~\ref{table:row-level-correlation} and Table~\ref{table:summary-level-correlation}.
    \item \textbf{Enhanced Throughput and Performance for Real-World Systems}: STRUM-LLM Distilled shows a 100-fold increase in throughput while being 10x smaller in size compared to models with similar performance. The system is capable of handling large amount of input text without being constrained to the context lengths supported by a given large language model. 
    \item \textbf{Critique-and-Revision Models}: STRUM-LLM builds task-specific critique-and-revision models in order to enhance the quality of data generation, and hence ensure the relevancy and accuracy of information. Table~\ref{table:CR} demonstrates that this approach improves the performance of STRUM-LLM Distilled by 14 points on our key metric -- the fraction of helpful rows in the output summary.
\end{itemize}

\section{Related Work}
There exists a line of work \citep{Roy2018TheoryAE,Angelidis2020ExtractiveOS,amplayo-etal-2021-aspect,ahuja-etal-2022-aspectnews} that focuses on aspect-based summarization of a single entity, which is not the goal of this work. Unlike single entity summarization methods, contrastive summarization literature is limited. \citet{Strhle2023ContrastiveTS} have surveyed contrastive text summarization methods, spanning from early rule-based systems to modern neural models. Among representative works, \citet{lerman-mcdonald-2009-contrastive} proposed an approach rooted in statistical language models that predominantly weighs the sentiments of opinions, and evaluated on consumer reviews. \citet{iso-etal-2022-comparative} introduced a neural technique for contrastive summarization, which they tested on a collection of short hotel reviews. Their principal contribution is a technique termed \textit{co-decoding}, which contrasts token probability distributions in contrastive summaries and aggregates them for shared summaries. However, unlike our work, their paper focuses on fully abstractive general (not attribute-based structured) contrastive summaries that optimize for token-level contrast, and most importantly limit the input source text to 16K tokens as their base model is a pre-trained LED model \cite{Beltagy2020LongformerTL}. The limit on input source length makes it unfair to compare to our method. Our approach builds on \cite{Gunel2023STRUMEA}, referred to as STRUM in-code in the rest of the paper, which generates extractive structured contrastive summaries using natural language inference and aspect extraction models. Although this approach can process arbitrarily long input sources, it is not efficient at inference time due to high number of pairwise calculations. This makes it hard to deploy in real-world applications, which was one of the main motivations of our method STRUM-LLM.


\section{Method}
We lay out the desiderata for our notion of a \textit{helpful comparison} between two \textit{entities}. We describe our STRUM-LLM approach and outline the critique-and-revision models that we build to improve the quality of our data generation pipeline.

\subsection{Desiderata for a \textit{Helpful Comparison}}
\label{sec:desiderata}

\begin{enumerate}
    \item \textbf{Attribution to Sources:} This ensures that the information presented is traceable, as each extraction is accompanied by evidence from the source text. We achieve this by adopting an extractive approach, and including source URLs for each extraction.
    \item \textbf{Identification of High-Contrast and Important Attributes:} We identify attributes where there is a significant contrast between the two options to aid with decision making. Note that some attributes are inherently informative and do not necessarily require contrasting views to be informative. We surface these attributes based on how often they occur in the input sources (popularity signal). Examples include the \textit{price} attribute for a car comparison or the \textit{ingredients} for a peanut butter comparison.
    \item \textbf{Consistent, Non-Redundant, and Accurate Opinion Representation:} We provide a comprehensive, consistent, and non-redundant view of the attributes and values. For example, if for entity A, the opinion \textit{good view} is held by a minority (1/10) and \textit{bad view} is the majority opinion (9/10), whereas for entity B, \textit{bad view} is unanimous (10/10), the comparison should not misleadingly present it as \textit{good view} for A versus \textit{bad view} for B (thereby optimizing contrast between the entities). We cluster the redundant values and remove the minority inconsistent values, so that the summary reflects the majority opinion.
    \item \textbf{Ranking and Presentation of Attributes:} This ensures that the most pertinent and high contrast attributes are highlighted in the comparison. We present the final comparison summary in a \textit{faceted} format where we show a row per attribute to further aid with structured decision making between two options.
\end{enumerate}

\subsection{STRUM-LLM}
\begin{figure}[h]
    \centering
    \includegraphics[width=\linewidth]{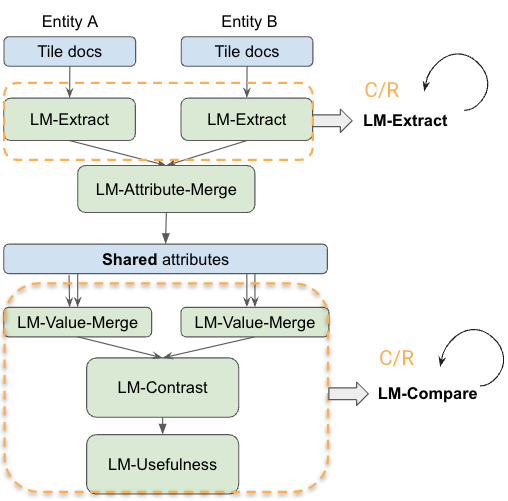}
    \caption{STRUM-LLM retrieves web pages relevant to the entities being compared, divides them into chunks of text that fit into the context window of the LLM that we refer to as \textit{tiling}, extracts attributes and values from the text, clusters related attributes and merges their values, and identifies the most meaningful contrast between the two entities. Critique-and-revision (CR) models improve the quality for both \textit{LM-Extract} and \textit{LM-Compare} during data generation.}
    \label{fig:strum-llm}
\end{figure}
STRUM-LLM pipeline is outlined in Figure~\ref{fig:strum-llm}. We use PaLM 2 \citep{Anil2023PaLM2T} as our base LLM models with a context window of 8K tokens across all LMs and critique-and-revision models. STRUM-LLM first retrieves 5-20 webpages from web search containing editorial text for both entities separately. Essential sentences from the webpages are extracted and tiled into chunks that can fit into the context window of the LLM.  There is no constraint on the total length of input sources fed into our pipeline due to our tiling approach.

\noindent \textbf{LM-Extract:} This LM extracts all the attributes and their corresponding values from the provided input text. We employ an extractive approach, drawing directly from the source text to ensure \textbf{attribution to sources}. Each extraction is accompanied by evidence from the source text including URLs, which we surface in the final structured summary, to maintain fidelity to the source. LM-Extract has several critique-and-revision models described in the next section, including \textit{insufficient context}, \textit{wrong entity}, and \textit{unhelpful attributes}. These critique-and-revision models ensure higher quality extractions while preserving the recall during data generation stage for distillation. LM-Extract is run for both entities separately.

\noindent \textbf{LM-Attribute-Merge:} This LM consolidates the attributes of both entities, merging similar attributes and only keeping the cluster centers to reduce redundancy among attributes.  This clustering stage helps with the \textbf{consistency} and the \textbf{redundancy} of the shown attributes.

\noindent \textbf{LM-Value-Merge:} This LM clusters the similar values for each attribute separately, removes inconsistent minority opinions, and keeps the cluster centers that reflect the majority opinions. This stage helps with the \textbf{consistency, redundancy, and accurate representation of opinions}. 

\noindent \textbf{LM-Contrast:} This stage identifies \textbf{attributes with contrasting values} and keeps important attributes that are inherently valuable for drawing insights. LM-Contrast also \textbf{ranks the attributes} based on their popularity and contrast level, taking the search ranking of the webpages of the extractions and the majority opinions into account.

\noindent \textbf{LM-Usefulness:} This LM filters comparison rows that are not useful based on the helpful comparison desiderata. We have this stage to catch the errors that might be made by any of the previous stages.

We refer to the combination of LM-Value-Merge, LM-Contrast, and LM-Usefulness run sequentially as \textbf{LM-Compare}. LM-Compare has several critique-and-revision models that we describe in the next section, including \textit{same attribute - orthogonal values}, \textit{inconsistent values}, \textit{unhelpful attributes}, \textit{over- and under-merged attributes}, and \textit{long complex claims}. These critique-and-revision models ensure higher quality comparisons in the data generation stage. We first generate data using few-shot prompted pre-trained LLMs for all the described LM stages above, and distill to a 10x smaller pre-trained LLM (STRUM-LLM Distilled) with a task mixture of LM-Extract ($\sim$15K row-level examples), LM-Attribute-Merge ($\sim$500 summary-level examples), and LM-Compare ($\sim$ 15K row-level examples) through fine-tuning on that generated data. We use that finetuned LM for \textbf{LM-Extract}, \textbf{LM-Attribute-Merge}, and \textbf{LM-Compare} in STRUM-LLM Distilled. Task mixture is the same for both when critique-and-revision models are on and when they are off. There is no human supervision at any stage of the data generation.

\subsection{Critique-and-Revision Models for STRUM-LLM}
The critique phase is where the LLM generates natural language critiques about its own outputs. Larger LLMs are often better at self-critiquing, despite dealing with more complex outputs \citep{Saunders2022SelfcritiquingMF}. Following the critique phase is the revision phase, where the LLM revises its output based on the critiques it generated. This involves improving the original output to enhance its accuracy, relevance, or coherence. In practice, critique-and-revision models can have significant implications in various applications. For instance, in the realm of creating harmless AI assistants, \citet{Bai2022ConstitutionalAH} employed self-critique and revision techniques during training phase following a list of rules they referred to as \textit{Constitutional AI}. 

We define a taxonomy of ways STRUM-LLM could produce an unhelpful summary based on our human evaluations and build targeted critique-and-revision (CR) models. We describe the individual CR models based on few-shot prompted pre-trained LLMs for \textit{LM-Extract} and \textit{LM-Compare} stages below. The critique model detects the described problem given the model output, and the revision model improves the output based on that problem while having access to the source text. We do not allow for additional retrieval of new webpages, but we allow for repeatedly invoking the CR models to iteratively refine the responses. The CR models are enabled for the data generation stage with the few-shot prompted LLMs to improve the finetuning data quality for the distilled model. \\

\noindent \textbf{CRs for \textit{LM-Extract}:}\\
\noindent \textbf{Insufficient Context:} There is insufficient context for the extracted value such as having \textit{views=good} while comparing hotels. Revision model repeats the extraction step using the source text for that attribute to provide more context such as \textit{views=good view from the rooftop, limited view from most of the rooms} for a hotel comparison.\\
\textbf{Wrong Entity:} Extractions associated with a different entity, such as including extractions related to \textit{iPhone 14 Pro} for \textit{iPhone 14}, are deleted. This error case happens when the entity name mention and the extraction are far apart in the text, hence appear in different tiles fed to the LM.\\
\textbf{Unhelpful Attributes:} The attributes extracted are not helpful such as \textit{color} for a \textit{winery}. The revision model deletes the unhelpful attributes.

\noindent \textbf{CRs for \textit{LM-Compare:}}\\
\textbf{Same Attribute - Orthogonal Values:} Values for the same attribute do not align across two entities such as showing \textit{fuel tank=iconic teardrop shape} for a motorcycle entity A and \textit{fuel tank=holds 3.3 gallons} for entity B. The revision model repeats the LM-Contrast stage to better align values.\\
\textbf{Inconsistent Values:} There are conflicting values (\textit{views=good, great, bad}) for the same entity. The revision model deletes the inconsistent values while preserving the majority opinion (\textit{views=good}).\\
\textbf{Unhelpful Attributes or Values:} Certain attributes or values might be unhelpful such as extracting \textit{wheels=four} for a car. The revision model removes the unhelpful attributes and values. \\
\textbf{Under-Merged or Over-Merged Attributes:} This CR fixes the issues with the attribute clustering which avoids showing redundant summaries.\\
\textbf{Long Complex Claims:} The LLM faces difficulties in determining contrast for long complex claims, so the revision model breaks these long claims into atomic claims.

\section{Evaluation Setup}
\begin{table*}[h]
\centering
\resizebox{\linewidth}{!}{
\begin{tabular}{|l|c|c|c|c|c|}
\hline
Model & \% of rows useful by CHS & Redundancy & Avg Inconsistent Values & Ranking Precision & Throughput \\
\hline
STRUM in-code \citep{Gunel2023STRUMEA} & 56.5\% & 19.6\% & 3.94 & 0.89 & 0.018 summ/s \\
STRUM-LLM Few-shot & 97.3\%  & 3.6\% & 0.89 & 0.91 & 0.013 summ/s\\
STRUM-LLM Distilled & 84.9\% & 4.0\% &1.26  & 0.92 & 1.500 summ/s\\
\hline
\end{tabular}}
\caption{We compare STRUM-LLM and STRUM-in-code models across comparison helpfulness score (CHS), summary-level metrics, and throughput. We observe the STRUM-LLM Distilled model outperforms STRUM in-code, and it is competitive with the STRUM-LLM Few-shot model that we use for data generation while having 100x higher throughput with a 10x smaller model.}
\label{table:main-table}
\end{table*}

\begin{table*}[h]
\centering
\resizebox{0.85\linewidth}{!}{
\begin{tabular}{|l|c|c|c|c|}
\hline
Model & \% of rows useful by CHS & Redundancy LLM  & Avg Inconsistent Values LLM & Ranking LLM \\
\hline
STRUM-LLM Distilled w/o CR & 79\% & 9.8\% & 1.69 & 0.97\\
STRUM-LLM Distilled w/ CR &  93\% & 2.5\% & 1.74 & 0.94\\
\hline
\end{tabular}}
\caption{Critique-and-Revision (CR) ablations across Shopping, Sports, Science, Home \& Garden categories including both LM-Extract and LM-Compare CRs. We observe that both comparison helpfulness score (CHS) and redundancy score improve without an impact on consistency and ranking precision.}
\label{table:CR}
\end{table*}
We design our evaluations to answer the following key questions: (1) Based on our \textit{helpful comparison} desiderata, does STRUM-LLM improve over the existing baseline in terms of quality and throughput? (2) Do our critique-and-revision models that enhance the data quality actually improve the performance of our distilled models? (3) Do our novel LLM-based automated metrics correlate to the human judgement? Note that LLM-based autoraters have been shown to have higher correlation to human judgements than traditional metrics such as ROUGE \citep{Lin2004ROUGEAP} or BERTScore \citep{Zhang2019BERTScoreET} in various natural language generation tasks based on their ability to capture task-specific properties \citep{Wang2023IsCA}. Hence, we build a few-shot prompted LLM-based \textit{comparison helpfulness scorer} (CHS) based on the desiderata we lay out in Section~\ref{sec:desiderata}, which operates on the row (attribute) level. We also develop few-shot prompted summary level pre-trained LLM-based autoraters that measure \textit{redundancy (clustering)}, \textit{consistency}, and \textit{ranking precision} in the final comparison summaries. We would like to note that the reported results are based on offline evaluation of the methods as described in this paper, and are not indicative of the deployed system end-to-end performance.

\subsection{STRUM In-Code Baseline}
Our pipeline builds on \citet{Gunel2023STRUMEA}, so we include it as an important baseline. STRUM in-code divides extracted essential sentences from the input webpages into chunks of 256 tokens. It uses a pre-trained large language model, fine-tuned on shopping-related data for high-precision attribute extraction \citep{Butane}, followed by agglomerative hierarchical clustering of attributes and values using a pre-trained natural language inference entailment model. For each shared attribute, source sentences are selected that maximize the contrast, based on the entailment model. Note that this approach is inherently slow due to high number of pairwise calls to the attribute discovery and entailment models. In addition, entailment models are brittle when used on sentences or on paragraphs that include complex claims, hence they should be used on propositions -- making the overall calculations even slower \citep{Chen2022PropSegmEntAL}. Overall, this approach is not suitable to deploy in real-world applications based on both quality and efficiency, evidenced by the measured throughput (number of summaries generated per second) in Table~\ref{table:main-table}.

\begin{table*}[h]
\centering
\resizebox{\linewidth}{!}{
\begin{tabular}{|l|c|c|c|c|c|}
\hline
Model & Avg \# of rows & \% of rows useful by Humans & \% of rows useful by CHS & Human Agreement & Human-CHS Agreement \\
\hline
STRUM in-code & 8.70 & 62\% & 70\% & 76\% & 83\% \\
STRUM-LLM & 10.12 & 81\% & 82\% & 88\% & 86\%\\
\hline
\end{tabular}}
\caption{Comparison helpfulness scorer (CHS) correlates well with human judgement. We also see that STRUM-LLM comfortably outperforms STRUM in-code on percentage of rows useful across summaries.}
\label{table:row-level-correlation}
\end{table*}

\begin{table*}[h]
\centering
\resizebox{\linewidth}{!}{
\begin{tabular}{|l|c|c|c|c|c|c|}
\hline
Model & Redundancy LLM & Redundancy Human & Inconsistency LLM  & Inconsistency Human & Ranking LLM & Ranking Human \\
\hline
STRUM in-code & 28.2\% & 23.7\%  & 2.74  & 2.23  & 0.88  & 0.73 \\
STRUM-LLM & 13.2\% & 9.8\% & 1.86  & 0.97 & 0.86 & 0.78  \\
\hline
\end{tabular}}
\caption{Summary-level LLM autoraters are well-correlated with human judgement. STRUM-LLM either outperforms or is comparable to STRUM in-code across all criteria.}
\label{table:summary-level-correlation}
\end{table*}

\subsection{Row-Level Comparison Helpfulness Evaluations}
Three human evaluators are provided with structured summaries comparing two entities where each row represents an attribute of comparison. Row-level ratings include \textit{YES, NO - Bad Extraction, NO - Inconsistent Values, NO - Undermerged Values, NO - Same Attribute, Orthogonal Values} that in detail in the Appendix.
In Table~\ref{table:row-level-correlation}, we correlate the comparison helpfulness scorer (CHS) to human ratings for both STRUM in-code and STRUM-LLM in order to demonstrate its credibility in determining a helpful comparison based on our desiderata. While estimating Human-CHS agreement, we take the majority opinion for humans, set the LLM's temperature to 0, and use the evaluations for our deployed system (total of a few hundred rows tested) for this analysis. Key metrics we measure for this row-level analysis are the average number of rows assessed, the fraction of rows marked useful by both humans and the LLM, and the levels of agreement among humans and between humans and the CHS. We observe a Human-CHS agreement either higher or comparable to the human agreement (Human-CHS Agreement of 83\% for STRUM in-code and 86\% for STRUM-LLM), confirming the trustworthiness of LLM-based evaluators at the row level. Also, in terms of \% rows useful by Humans, we demonstrate that STRUM-LLM (81\%) outperforms STRUM in-code (62\%).

\subsection{Summary-Level Evaluations}
\noindent \textbf{Redundancy (Clustering):} We measure attribute clustering efficacy with this metric. As an example, \textit{room} and \textit{rooms} or \textit{amenities} and \textit{facilities} should not be separate rows in the structured summary. More formally, we calculate this metric as $1 - \frac{\text{number of unique attribute clusters}}{\text{number of all attributes}}$ per summary.\\
\textbf{Consistency:} We check for inconsistent values in the summary across different attributes. We define this metric by the raw \textit{number of inconsistent values}. We do not calculate the ratio of inconsistent values for better granularity as the percentage is very low for both STRUM-LLM and STRUM in-code. As an example, for the set of values [45 liters, 46 liters, 46 liters in volume, 46 liters of space], the number of inconsistent values is 1 as one would need to delete the inconsistent value of 45 liters to make all the other values consistent. \\
\textbf{Ranking Precision:} The output structured summary table should prioritize rows that are helpful to the comparison and rank them accordingly. The concrete metric we calculate is \textit{Precision at 5}.

In Table~\ref{table:summary-level-correlation}, we demonstrate that summary-level LLM autoraters (redundancy, consistency, and ranking precision), when temperature is set to 0, are well-correlated to the human judgement. Finally, we show that STRUM-LLM either outperforms or is comparable to STRUM in-code across all summary-level criteria. Notably, STRUM-LLM decreases the redundancy to 9.8\% from 23.7\% of STRUM in-code based on human evaluations.

\section{Results}

In Table~\ref{table:main-table}, we present our main results using the row-level and the summary-level metrics. Our test set consists of 250 sampled A vs B queries across the categories: Health, Computers \& Electronics, Arts \& Entertainment, Jobs \& Education, Autos \& Vehicles, Travel \& Transportation, Food \& Drink, Shopping, Sports, Science, Beauty \& Fitness, People \& Society, Finance, Pets \& Animals, Games, News, Internet \& Telecom, Home \& Garden, and Business \& Industrial.

STRUM-LLM Distilled significantly outperforms STRUM-in-code across all metrics while having 100x more throughput. Particularly notable was its performance in the CHS and redundancy metrics: STRUM-LLM Distilled has 84.9\% rows marked useful by CHS and 4\% redundancy across the summary while STRUM in-code has 56.5\% rows marked useful by CHS and 19.6\% redundancy across the summary. Also, STRUM-LLM Distilled model is remarkably competitive with the STRUM-LLM Few-shot model across all summary-level metrics. This is in stark contrast to the STRUM-in-code approach, which is notably slower.  Note that STRUM-LLM Few-shot model's \% of rows useful by CHS is expected to be close to 100\% as it has an LM-Usefulness step in the end for higher data quality. Finally, In Table~\ref{table:CR}, we evaluate how effective the CR models are for both LM-Extract and LM Compare for the STRUM-LLM Distilled model across Shopping, Sports, Science, Home \& Garden categories. Most notably, \% of rows useful by CHS goes from 79\% to 93\% and redundancy decreases to 2.5\% from 9.8\% without an impact on consistency and ranking precision when CRs are on. It is important to note that building new CR models allow us to improve the data quality whenever we detect a problem trend in our evaluations for our deployed system. We include several comparison examples from our pipeline in the Appendix.

\section{Conclusion and Future Work}
We introduce STRUM-LLM -- a novel domain-agnostic approach for generating attributed and structured contrastive summaries that facilitate informed decision-making between two choices. Our method does not require any human-labeled data or pre-determined attribute list as supervision, and it does not have a limit on the length of input sources it can process. Importantly, STRUM-LLM adheres to the desiderata we lay out for a helpful comparison, namely: attribution to sources; identification of high-contrast and important attributes; consistent, non-redundant, and accurate opinion representation, and prioritization of relevant attributes. We demonstrate that critique-and-revision models improve the quality. Through our extensive evaluations, we show that STRUM-LLM produces high quality comparison summaries and that its distilled version has 100x more throughput than the models with comparable performance while being 10x smaller. For future work, we aim to accommodate a wider variety of query types beyond pairwise comparisons and integrate multimodal data such as images and figures in addition to text.

\section{Ethical Considerations}
We acknowledge that our STRUM-LLM approach is limited by the factuality of the top web search results. Hence, if a webpage presents an outdated or an incorrect piece of information, this might surface in our final comparison summary. In addition, we acknowledge that the notion of an \textit{important attribute} for an A vs. B comparison is highly personal and subjective, hence might be different for different groups. We make the assumption that popular and contrastive attributes are important for A vs. B decision making.
\bibliography{anthology,custom}

\appendix
\label{sec:appendix}

\section{Row-Level Comparison Helpfulness Evaluations}
Three human evaluators are provided with structured summaries comparing two entities where each row represents an attribute of comparison. We describe the predefined row-level ratings below. \\\\
\noindent \textbf{YES:} Both the attribute and shown values are related to the comparison and satisfy the desiderata laid out in Section~\ref{sec:desiderata}. \\
\textbf{NO - Bad Extraction:} Either attribute or the value extraction does not make sense such as extracting \textit{graphic card} for a campsite or extracting \textit{bright} for the \textit{transmission} attribute in a car.\\
\textbf{NO - Inconsistent Values:} The values for a single entity are inconsistent. An example would be having \textit{monthly fee=$\$10$, $\$5000$} or \textit{price=expensive, affordable} for the same entity.\\
\textbf{NO - Undermerged Values:} Some values are redundant and should be merged with the other values such as \textit{backpack} and \textit{travel backpack}.\\
\textbf{NO - Same Attribute, Orthogonal Values:} Values of the same attribute are interpreted differently across the two entities. An example would be having \textit{steering=sporty and fun to drive} for one entity and \textit{accurate} for the other one.\\
\textbf{OK:} Row does not fit any of the NO categories but it does fit the helpful comparison desiderata either.

\section{STRUM-LLM Output Summaries}
We show three STRUM-LLM summaries that are output by our pipeline. Note that each shown extraction is attributable to the source, and includes \textit{other sources} or \textit{conflicts} tags if other sources support or conflict with the shown extractions. 
\begin{figure*}[h]
\vspace{-1cm}
    \centering
    \includegraphics[width=0.95\linewidth]{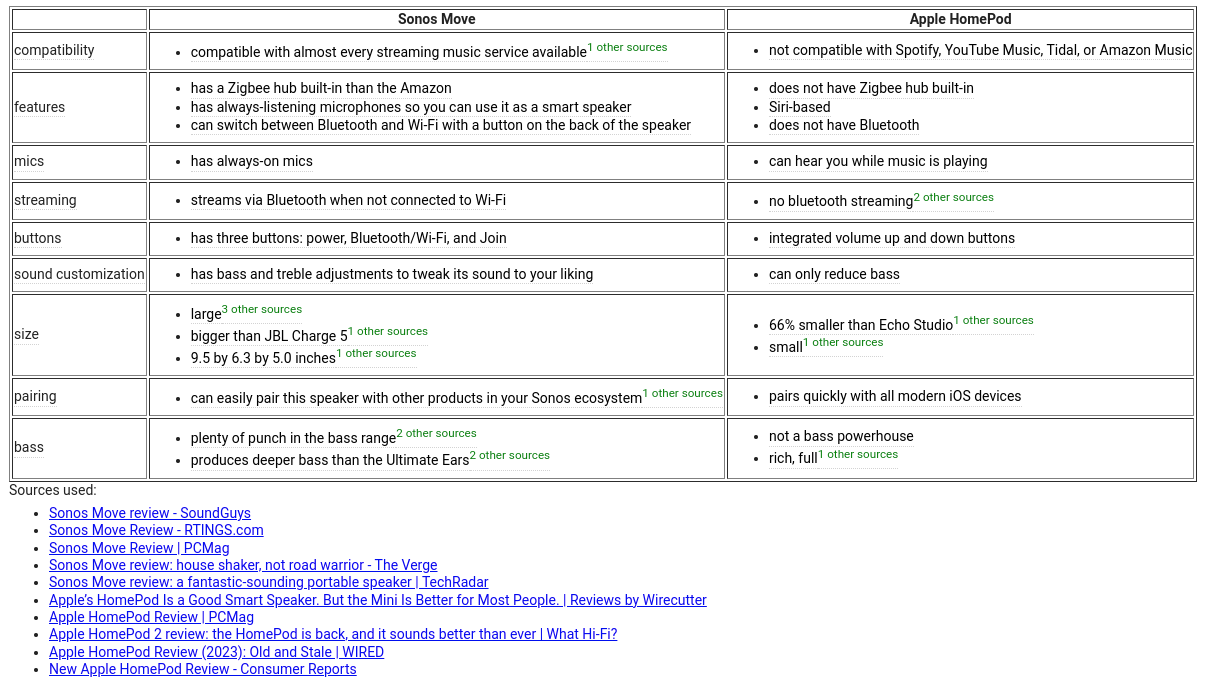}
    \caption{STRUM-LLM summary comparing \textit{Sonos Move} and \textit{Apple HomePod}.}
    \label{fig:strum-ex1}
\vspace{-1cm}
\end{figure*}
\begin{figure*}[h]
    \centering
    \includegraphics[width=0.95\linewidth]{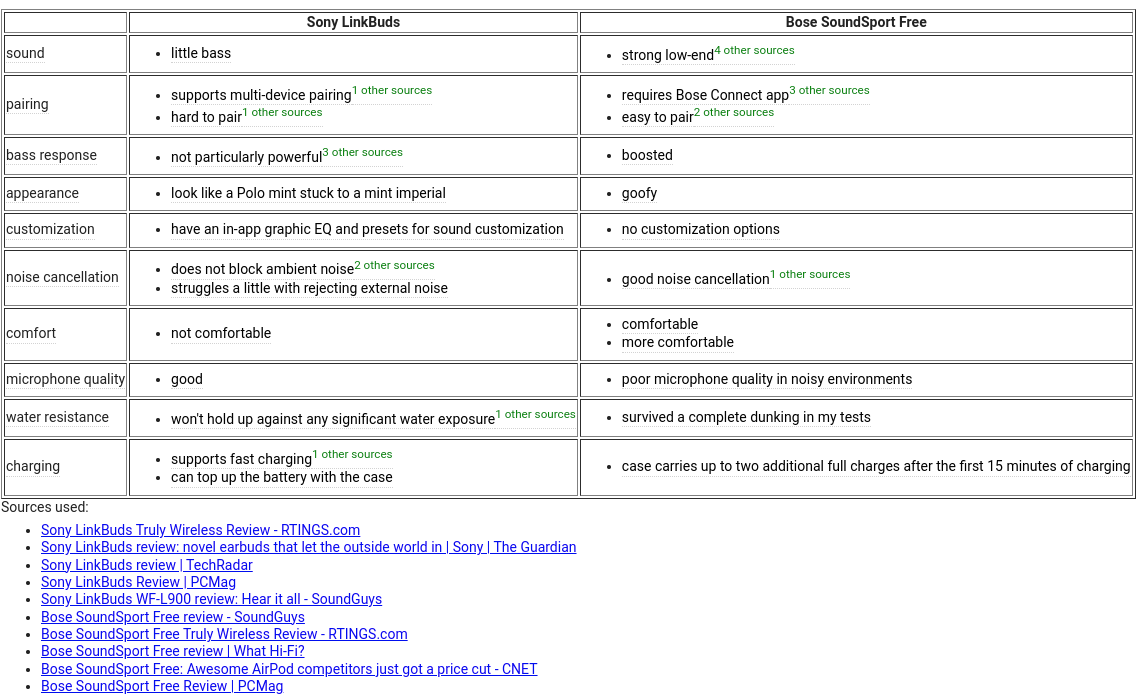}
    \caption{STRUM-LLM summary comparing \textit{Sony LinkBuds} and \textit{Bose SoundSport Free}.}
    \label{fig:strum-ex2}
\end{figure*}
\begin{figure*}[h]
    \centering
    \includegraphics[width=\linewidth]{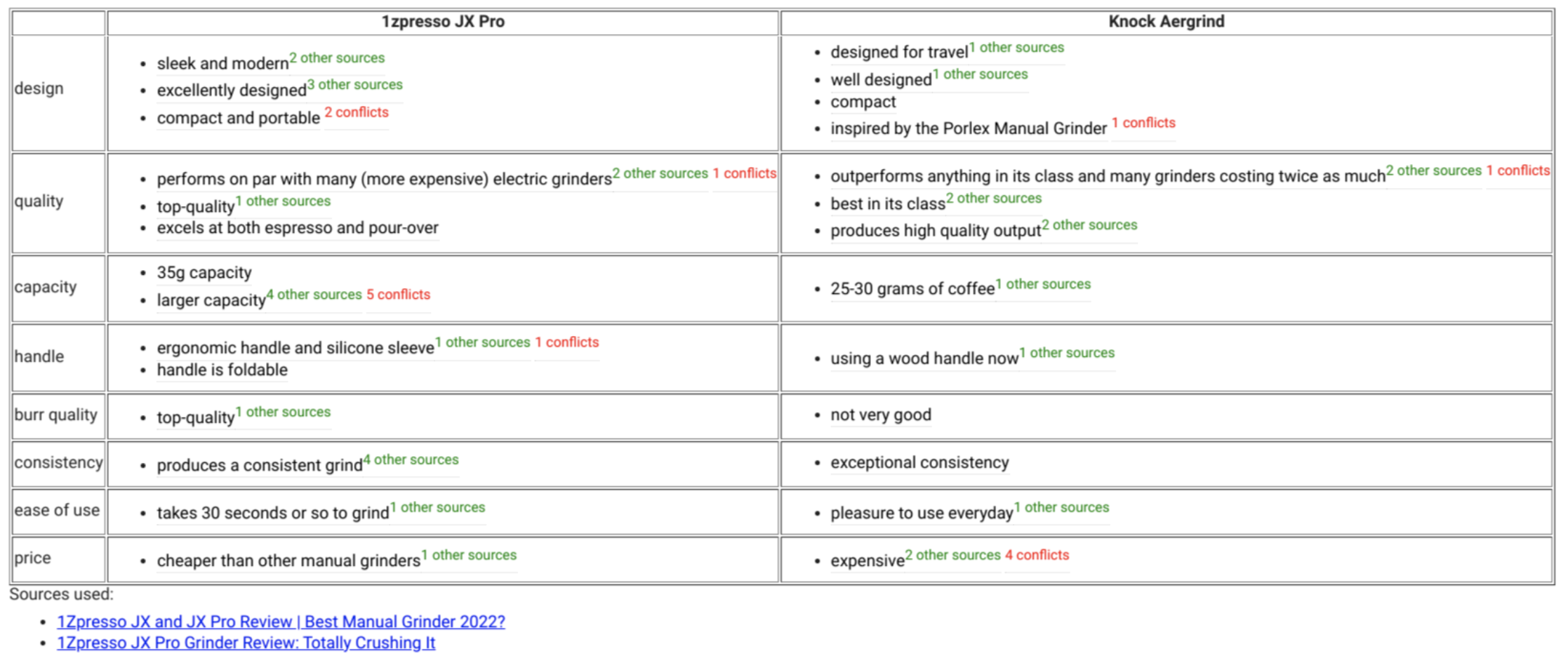}
    \caption{STRUM-LLM summary comparing \textit{1zpresso JX Pro} and \textit{Knock Aergrind}.}
    \label{fig:strum-ex3}
\vspace{-1cm}
\end{figure*}

\begin{figure*}[h]
    \centering
    \label{fig:strum-ex4}
\end{figure*}

\end{document}